\documentclass[conference]{IEEEtran}

\usepackage{cite}
\usepackage{amsmath,amssymb,amsfonts}
\usepackage{algorithmicx}
\usepackage{graphicx}
\usepackage{textcomp}
\usepackage{xcolor}
\usepackage{hyperref}

\AtBeginDocument{%
  \providecommand\BibTeX{{%
    \normalfont B\kern-0.5em{\scshape i\kern-0.25em b}\kern-0.8em\TeX}}}

\usepackage{graphicx}
\usepackage{amsmath}
\usepackage{amssymb}
\usepackage{booktabs}
\usepackage{rotating}
\usepackage{xfrac}

\usepackage{hyperref}

\usepackage[acronym]{glossaries}
\makeglossaries
\newacronym{dnns}{DNN's}{Deep Neural Network's}

\newacronym{fa}{$\mathtt{FedAvg}$}{Federated Averaging}

\newacronym{fl}{FL}{Federated Learning}

\newacronym{ipa}{IPA}{Iterative Parameter Alignment}

\usepackage{times}
\usepackage{epsfig}
\usepackage{comment}
\usepackage{graphicx}
\usepackage{amsmath}
\usepackage{amssymb}
\usepackage{booktabs}
\usepackage{makecell}
\usepackage{multirow}
\usepackage{bm}
\usepackage{arydshln}

\usepackage[utf8]{inputenc} % allow utf-8 input
\usepackage[T1]{fontenc}    % use 8-bit T1 fonts
\usepackage{hyperref}       % hyperlinks
\usepackage{url}            % simple URL typesetting
\usepackage{booktabs}       % professional-quality tables
\usepackage{amsfonts}       % blackboard math symbols
\usepackage{nicefrac}       % compact symbols for 1/2, etc.
\usepackage{microtype}      % microtypography

\usepackage{amssymb}

\usepackage[acronym]{glossaries}
\usepackage{comment}
\usepackage{algorithm}
\usepackage{algpseudocode}
\usepackage{wrapfig}
\usepackage{booktabs}
\usepackage{tabularx}
\usepackage{makecell}

\usepackage{amsmath}

\usepackage{caption}
\usepackage{subcaption}
\usepackage{caption,stackengine}
\usepackage{enumitem}

\usepackage{capt-of}% or \usepackage{caption}
\usepackage{booktabs}
\usepackage{varwidth}
\usepackage{framed}
\DeclareMathOperator*{\argmin}{arg\,min}

\usepackage{xpatch}

\makeatletter
\xpatchcmd{\algorithmic}
  {\ALG@tlm\z@}{\leftmargin\z@\ALG@tlm\z@}
  {}{}
\makeatother

\algnewcommand\algorithmicforeach{\textbf{for each} }
\algdef{S}[FOR]{ForEach}[1]{\algorithmicforeach\ #1\ \algorithmicdo}
\algrenewcommand\algorithmicfunction{}
\algtext*{EndFunction}
\algtext*{EndFor}

\algdef{SE}[SUBALG]{Indent}{EndIndent}{}{\algorithmicend\ }%
\algtext*{Indent}
\algtext*{EndIndent}

\usepackage{amsmath,amssymb}
\makeatletter
\newsavebox\myboxA
\newsavebox\myboxB
\newlength\mylenA

\newcommand*\xoverline[2][0.75]{%
    \sbox{\myboxA}{$\m@th#2$}%
    \setbox\myboxB\null% Phantom box
    \ht\myboxB=\ht\myboxA%
    \dp\myboxB=\dp\myboxA%
    \wd\myboxB=#1\wd\myboxA% Scale phantom
    \sbox\myboxB{$\m@th\overline{\copy\myboxB}$}%  Overlined phantom
    \setlength\mylenA{\the\wd\myboxA}%   calc width diff
    \addtolength\mylenA{-\the\wd\myboxB}%
    \ifdim\wd\myboxB<\wd\myboxA%
       \rlap{\hskip 0.5\mylenA\usebox\myboxB}{\usebox\myboxA}%
    \else
        \hskip -0.5\mylenA\rlap{\usebox\myboxA}{\hskip 0.5\mylenA\usebox\myboxB}%
    \fi}
\makeatother

\newcommand{\tolstrut}{%
  \vrule height\dimexpr\fontcharht\font`\A+.1ex\relax width 0pt\relax
}

\DeclareRobustCommand{\textoverline}[1]{%
  \ensuremath{\overline{\mbox{\tolstrut#1}}}%
}

\usepackage{calc}
\usepackage{hyperref}
\hypersetup{colorlinks=true}

\makeatletter
\newcommand*\@dblLabelI {}
\newcommand*\@dblLabelII {}
\newcommand*\@dblequationAux {}

\def\@dblequationAux #1,#2,%
    {\def\@dblLabelI{\label{#1}}\def\@dblLabelII{\label{#2}}}

\newcommand*{\doubleequation}[3][]{%
    \par\vskip\abovedisplayskip\noindent
    \if\relax\detokenize{#1}\relax
       \let\@dblLabelI\@empty
       \let\@dblLabelII\@empty
    \else % we assume here that the optional argument
       \@dblequationAux #1,%
    \fi
    \makebox[0.5\linewidth-1em]{%
     \hspace{\stretch2}%
     \makebox[0pt]{$\displaystyle #2$}%
     \hspace{\stretch1}%
    }%
    \makebox[0.5\linewidth-1em]{%
     \hspace{\stretch1}%
     \makebox[0pt]{$\displaystyle #3$}%
     \hspace{\stretch2}%
    }%
    \makebox[2em][r]{(%
  \refstepcounter{equation}\theequation\@dblLabelI, 
  \refstepcounter{equation}\theequation\@dblLabelII)}%
  \par\vskip\belowdisplayskip
}
\makeatother

\begin{document}

%\title{Cross-Domain Federated Learning with Iterative Parameter Alignment}
\title{Cross-Silo Federated Learning Across Divergent Domains with Iterative Parameter Alignment}

%\title{Collaborative Learning Across Divergent Domains with Iterative Parameter Alignment}


\author{\IEEEauthorblockN{Matt Gorbett}
\IEEEauthorblockA{\textit{Department of Computer Science} \\
\textit{Colorado State University}\\
Fort Collins, CO, United States \\
Matt.Gorbett@colostate.edu}
\and
\IEEEauthorblockN{Hossein Shirazi}
\IEEEauthorblockA{\textit{Fowler College of Business} \\
\textit{San Diego State University }\\
San Diego, CA, United States \\
hshirazi@sdsu.edu}
\and
\IEEEauthorblockN{Indrakshi Ray}
\IEEEauthorblockA{\textit{Department of Computer Science} \\
\textit{Colorado State University}\\
Fort Collins, CO, United States \\
Indrakshi.Ray@colostate.edu}

}

\IEEEoverridecommandlockouts
\IEEEpubid{\makebox[\columnwidth]{979-8-3503-2445-7/23/\$31.00~\copyright2023 IEEE \hfill}
\hspace{\columnsep}\makebox[\columnwidth]{ }}

\maketitle

\IEEEpubidadjcol

\begin{abstract}

%Learning from the collective knowledge of data dispersed across private sources can provide neural networks with enhanced generalization capabilities.
%Federated learning collaboratively trains a machine learning model across remote sources by combining client models via the orchestration of a central server. 

Learning from the collective knowledge of data dispersed across private sources can provide neural networks with enhanced generalization capabilities.
Federated learning, a method for collaboratively training a machine learning model across remote clients, achieves this by combining client models  via the orchestration of a central server. 
However, current approaches face two critical limitations: i) they struggle to converge when client domains are sufficiently different, and ii) current aggregation techniques produce an identical global model for each client. 
In this work, we address these issues by reformulating the typical federated learning setup: rather than learning a single global model, we learn $\mathnormal{N}$ models each optimized for a common objective.  
To achieve this, we apply a weighted distance minimization to model parameters shared in a peer-to-peer topology.
The resulting framework, Iterative Parameter Alignment, applies naturally to the cross-silo setting, and has the following properties: (i) a unique solution for each participant, with the option to globally converge each model in the federation, and (ii) an optional early-stopping mechanism to elicit \textit{fairness} among peers in collaborative learning settings.
These characteristics jointly provide a flexible new framework for iteratively learning from peer models trained on disparate datasets. 
We find that the technique achieves competitive results  on a variety of data partitions compared to state-of-the-art approaches. 
Further, we show that the method is robust to divergent domains (i.e. disjoint classes across peers) where existing approaches struggle.

 %The method, Iterative Parameter Alignment, learns a global 

%%Federated learning, the predominant method for training across decentralized data sources, 

%Learning from the collective knowledge of data dispersed across private devices can provide neural networks with enhanced generalization capabilities. Federated learning has emerged as promising 

%Federated learning, the predominant method for achieving this, collaboratively trains a shared model across decentralized data sources.  Current solutions aggregate client models trained on local datasets into a cohesive global model via the orchestration of a central server. However, global models in such frameworks suffer from suboptimal solutions due to non-identical client data distributions, while local models do not benefit from the knowledge of peers.

%To achieve this, we share model parameters across peers, aligning them with a simple addition to the objective function. 

%Federated 

%suffers from sub-optimal solutions due to imbalanced device data partitioning. In this work, we reframe the typical federated learning framework: instead of learning a single global model, we learn N globally optimized local models. To achieve this, we share model parameters across peers, aligning them with a simple addition to the objective function. 
\end{abstract}

\section{Introduction}\label{intro}

\quad \gls{fl} addresses issues of data privacy and access rights by enabling wide-scale training of machine learning models across decentralized data sources \cite{li2020federated,mcmahan2017communication,yang2019federated}.
Standard \gls{fl} involves clients (e.g. mobile, edge devices) training a model locally with private data and communicating their model updates back to a central server.
The server aggregates client models into a global model and returns it to each client, %, repeating this process until convergence.  
a process that repeats iteratively until a final global model is produced.  
Traditional \gls{fl} often addresses \textit{cross-device} settings where clients consist of unreliable devices. Extensive research has concentrated on addressing issues related to cross-device \gls{fl} such as communication constraints and heterogeneous data partitioning \cite{guo2021hybrid,kairouz2021advances,acar2021federated,gao2022feddc,li2020federated,karimireddy2020scaffold}.

A second setting, \textit{cross-silo} \gls{fl}, involves training a machine learning model across large organizations such as banks \cite{chase_fed, editor2fedai_webank_nodate} and hospitals \cite{dayan2021federated,ogier_du_terrail_federated_2023,silva_federated_2019,flores_nvidia_2020}. 
Silos in this scenario generally have big data, extensive computational resources, and strong network communication \cite{huang2022cross}.
Further, the setting often contains fewer clients compared to cross-device \gls{fl}. %Clients in cross-silo \gls{fl} often have as much capacity as a central orchestrator;

% in fact, the central server may in fact inhibit the speed of learning due to a communication bottleneck.  A natural direction for cross-silo \gls{fl} is to then replace the client-server architecture with peer-to-peer communication \cite{pmlr-v84-bellet18a, pmlr-v48-colin16,koloskova2019decentralized,zhao2022beer,wang2019matcha}, a topology that was studied in depth by Marfoq et al.  \cite{marfoq2020throughput}. %In this work, we concentrate on the cross-silo \gls{fl} setting.  

\begin{figure*}[t]
\includegraphics[width=1\textwidth]{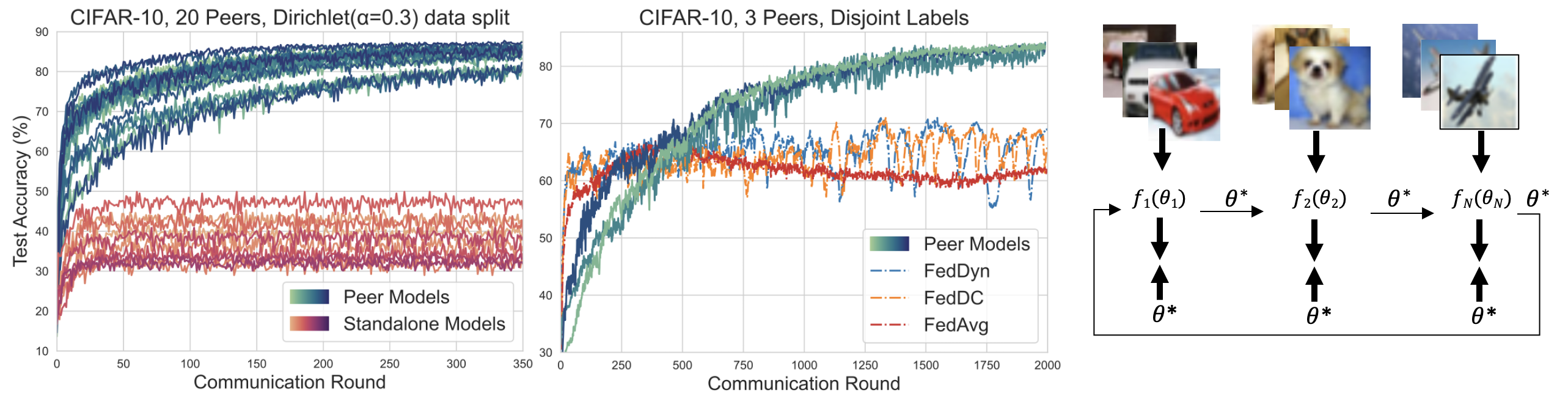}
\caption{\textbf{Left: } Test set accuracy across communication rounds of peers trained with Iterative Parameter Alignment compared to their standalone performance (trained only on their local data).  There are twenty peers each trained with an imbalanced subset of the CIFAR-10 training set.  They are split using heterogeneous data partitioning using a Dirichlet distribution with $\alpha=0.3$. One communication round (the x-axis) equals each $peer_i$ training their model ($f_i$) once. \textbf{Center:} Three peers each trained with distinct CIFAR-10 training labels (one peer has 4 labels, two peers have 3 labels each).  We find that when peers have sufficiently divergent domains, existing \gls{fl} methods fail, creating global models that do not reach baseline accuracy.  Iterative Parameter Alignment produces distinct global models for each peer that each converge to baseline accuracy (85\% on the test set). \textbf{Right:} A single iteration of Parameter Alignment trained in a ring topology (random topologies are used in experiments).  The method relies on parameter exchange and alignment to learn from others. $\theta_1,\theta_2,...\theta_N$ are $N$ peers parameters and $f_1,f_2,... f_N$ are the models.  $\theta^*$ represents all peer parameters $\{\theta_1,\theta_2,...\theta_N\}$. Each $peer_i$ can optionally apply differential privacy to their $\theta_i$ for protection. Our code is available at \url{https://github.com/mattgorb/iterative_parameter_alignment}.}
\label{fig:arc}
\end{figure*}

\textbf{Motivation.} In this work we identify and address two issues present in current \gls{fl} algorithms.  First,  we identify a novel failure scenario in current \gls{fl} frameworks: cross-domain global model aggregation. Specifically, when clients have divergent domains, such as completely different labels, common \gls{fl} approaches fail.  
Figure \ref{fig:arc} (center) highlights the issue, with existing algorithms FedAvg \cite{mcmahan2017communication}, FedDC \cite{gao2022feddc}, and FedDyn \cite{acar2021federated} each failing to converge to baseline test accuracy when three clients have differing labels (e.g. client one has only training samples of animals and another only vehicles). %In this scenario, two clients have three CIFAR-10 training labels and one client has four labels.  
Cross-domain scenarios are important in the real-world such as
those involving GDPR where an entire demographic segment
is isolated, or cross-industry learning where the domains of peers are disjoint.
In Section \ref{segregate}, we show that existing \gls{fl} algorithms consistently have unstable results across various datasets and label splits.  %We relate this issue to domain divergence, an active area of research in negative transfer learning \cite{wang2019characterizing,zhang2022survey}, where source domain knowledge negatively effects a target domain's ability to learn.%Divergent domains is also an issue in transfer learning when negative transfer occurs as a result of the source and target domains not being sufficiently similar. 

In addition, we also address an overlooked characteristic of existing \gls{fl}: \textit{the global model is identical for each participant}. 
This property can lead to important disadvantages.
In particular, the global model is exposed to all participants in the federation. In the cross-silo setting this may leave a client model unprotected against direct competitors, exposing obvious vulnerabilities such as white-box attacks \cite{goodfellow2014explaining}. Personalized \gls{fl} is an alternative approach which produces individualized models for each client unique to their data distribution \cite{fallah2020personalized,zhang2020personalized,li2021ditto}, including methods to produce joint personalized and global models \cite{chen2021bridging, chen2023the}, however current approaches still produce a \textit{single} global model (details in Section \ref{related}).  
%Additionally, the client-server architecture exposes the system to a single point of failure at the server.  

\textbf{Proposed Approach.} To address the issues of divergent client domains and a single global model, we propose the \gls{ipa} algorithm
for merging machine learning models across silos. Unique from existing approaches the algorithm trains $N$ different models, one for each silo.  The models each have arbitrary initializations, different from current techniques which require the same initial parameters \cite{mcmahan2017communication}.  \gls{ipa} works by iteratively merging the models by minimizing the distance between weights.  The architecture  is depicted in Figure \ref{fig:arc} (right).  
%\gls{ipa} addresses the
%\gls{ipa} offers several advantages compared to existing methods:

%\vspace{-2.5em}

%Essential to cross-silo \gls{fl},
%\gls{ipa} addresses security of both the peers data \textit{and} its final model.  
Essential to cross-silo \gls{fl}, \gls{ipa} can protect the 
client's data \textit{and} the client's final model. Data protection is a primary goal of \gls{fl} (achieved via data localization \cite{mcmahan2017communication} and differential privacy \cite{liu2022on,geyer2017differentially,agarwal2021skellam,kairouz2021distributed}), however, model protection is a more ambiguous task.  Homomorphic encryption is the primary model protection technique in \gls{fl}, enabling clients to encrypt model updates for protection against a central server \cite{zhang2020batchcrypt, jiang2021flashe}. Differential privacy also achieves model protection by enabling clients to add noise to their models parameters \cite{wei2020federated}.  However, in each of these scenarios the global model is still the same for each client.  Some techniques emphasize fairness by producing varying models which depend on a clients data contribution \cite{xu2021gradient,lyu2020towards,lyu2020collaborative}. However, such approaches produce client models derived from the same global model, which are produced at the server. Moreover, fairness may not be a requirement for every \gls{fl} scenario, and in such cases it may still be desirable to create differing global models without this constraint.  

\gls{ipa} addresses issues of model protection through silo-to-silo federated learning over $N$ unique models. This decentralized topology is favorable particularly in cross-silo scenarios where a central server may create a communication bottleneck \cite{marfoq2020throughput}. By creating unique global models for each peer, \gls{ipa} prevents peers from knowing each others parameters, an approach achieved through differential privacy.  In particular, a peer may add noise to its model parameters to protect its model from the other peers (we assess \gls{ipa} under differential privacy in Section \ref{abl}).  We study the differences in peer models without differential privacy in Section \ref{comparison}. 

In addition to model protection, \gls{ipa} provides the flexibility for peer models to either converge to a global optimum, or decide on an early-stopping point to elicit fairness (in cases of heterogeneous data silos). For example, when one peer has more data than another, their model will converge faster using the \gls{ipa} algorithm. If fairness is a requirement, peers can decide on an early stopping point so that the higher contributor achieves a stronger model.  If fairness is not a requirement, all peers will still converge to a global optimum (each with unique parameters).   
We study this property of \gls{ipa} in Section \ref{fairness_section}.

\textbf{Contributions.} We propose \gls{ipa}
for merging peer models trained on separate data. Different from existing approaches, the algorithm produces a unique model for each peer (silo) in the federation, each with arbitrary initializations.  \gls{ipa} works by iteratively merging peer models by minimizing the distance between weights.  The architecture  is depicted in Figure \ref{fig:arc} (right).  \gls{ipa} offers several advantages compared to existing methods:

\begin{itemize}
    \item \gls{ipa} is robust in scenarios with completely segregated labels across peers, including scenarios where existing \gls{fl} algorithms fail to converge. 
    \item  \gls{ipa} achieves state-of-the-art convergence rates on balanced data partitions (Table \ref{statistics}).  Further, the method achieves competitive results (significantly outperforming FedAvg) with heterogeneous data sources, a known burden of standard \gls{fl} \cite{mcmahan2017communication,acar2021federated,gao2022feddc,li2020federated,karimireddy2020scaffold}. 
        \item It produces unique peer models in a decentralized topology, providing independence from a central orchestrator and implicit collaboration with peers. 
    \item  The method produces distinct global models for each peer, which we analyze in Section \ref{comparison}.
    \item \gls{ipa} contains \textit{built-in} fairness: we show that model performance on classification tasks is correlated with a peers standalone model performance. We propose an early stopping mechanism to elicit fairness in Section \ref{fairness_section}.  
\end{itemize}

\section{Related Work } \label{related}
%Federated learning has produced an extensive library of research relating our work to various subfields of \gls{fl}
%We provide a comprehensive related work in the Appendix.  

\textbf{Federated Learning.}   The pioneering \gls{fl} framework, \gls{fa},  aggregated a global model by averaging the weights of client models trained on private data \cite{mcmahan2017communication}; heterogeneous data partitioning, inefficient communication, and variable participation  across clients were identified as key challenges \cite{khaled2020tighter, woodworth2018graph, konevcny2016federated}.  Subsequent work improved the convergence rate of heterogeneous client data through corrections to the gradients of local models \cite{karimireddy2020scaffold}, regularization of local models against the global model \cite{li2020federated}, dynamic regularization of local models \cite{acar2021federated}, and correcting local model drift from the global model \cite{gao2022feddc}.

\textbf{Cross-Silo Federated Learning.} Cross-silo \gls{fl} involves training machine learning models across entities with large data-silos (i.e. data centers) \cite{kairouz2021advances,huang2022cross}. Peer-to-peer communication has been proposed as an effective alternative to centralized orchestration in cross-silo federations with reliable participants \cite{marfoq2020throughput}.  Marfoq et al. examine the effect of topology on the duration of communication rounds in cross-silo
settings, and propose algorithms for measuring network characteristics to construct a high-throughput network topology. 
%Guo et al. \cite{guo2021hybrid} use a hybrid device-to-device and device-to-server framework to improve communication in heterogeneous \gls{fl} settings.  
Other works address security and personalization of cross-silo \gls{fl} \cite{liu2022on,heikkila2020differentially}.
We consider cross-silo \gls{fl} a realistic application for \gls{ipa} due to the large computational costs of the algorithm, as well as organizations' potential desire to maintain independent models.  

\textbf{Collaborative Learning.}
Important to cross-silo \gls{fl} is designing incentive mechanisms for peers to participate in a federation, commonly referred to as collaborative learning. 
Participants may have concerns about contributing their data for the benefit of others. For example, if two peers are direct competitors they may be concerned that the other peer will benefit more from federated learning.   As a result, \textit{fairness} schemes have been proposed using methods such as contract theory \cite{8832210,kang2019incentive}, monetary payouts \cite{yu2020sustainable}, and game-theoretic approaches \cite{donahue2021optimality, blum2021one}.  Lyu et al. \cite{lyu2020towards} propose a credibility metric so that each participant receives a different version of the global model with performance comparable to its contribution.  Similar to our work the authors use a decentralized framework (they propose blockchain). Different from their approach, \gls{ipa} works in cross-domain settings and produces differing global models.  \gls{ipa} is additionally a less complex framework.  
Xu et al.  \cite{xu2021gradient} propose a reward mechanism that specifies model updates at the server commensurate to a client's contributions.  Other works utilize the Shapely value \cite{10.1145/3501811} and reputation lists \cite{lyu2020collaborative} to evaluate client contributions.

\textbf{Personalized Federated Learning.}\label{personalized_related}  Personalized \gls{fl} produces individualized models that are catered to a client's data distribution while also leveraging the data of the federation \cite{fallah2020personalized}.  
Clients can create personalized models via local fine-tuning  of the global model \cite{kairouz2021advances}, or from more advanced techniques such as hypernetworks \cite{shamsian2021personalized}, pruning \cite{9545941}, encouraging interaction between related clients \cite{smith2017federated, zhang2020personalized, chen2023the,huang2021personalized}, and learning client-level and shared feature extractors \cite{liang2020think,collins2021exploiting}.  Research also addresses \textit{fairness} in personalized \gls{fl} \cite{mohri2019agnostic, li2021ditto}, identifying performance disparity across clients as a key issue.  

Some methods create high performing personalized \textit{and} global models.  \texttt{FedRoD} \cite{chen2021bridging} utilizes an additional local layer on a global model to create a high performing personalized model, while \texttt{FedHKD} \cite{chen2023the} uses local "hyper knowledge" to aggregate the global model.  However, these approaches create identical global models across clients.  Further, the methods centrally aggregate the global model.  

\textit{\gls{ipa} versus personalized \gls{fl}.
}
%\gls{ipa} instead creates an individualized model for each peer which can perform well on a global task, while merging peer knowledge from completely independent domains. 
Unique from \gls{ipa}, personalized \gls{fl} methods produce models \textit{individualized to each clients data distribution}. For example, if one client has data of dogs and another has data of cats, they may not benefit from each other. Unlike existing research in personalized \gls{fl}, \gls{ipa} aims to learn an individualized model on a common objective for each peer in the network.  In our dogs and cats example, each peer would learn a different global model that does well at classifying dogs and cats in a decentralized network topology.

\section{Method}

We begin by reviewing the standard federated averaging objective, followed by describing the unique approach of \gls{ipa}.  

%We begin by reviewing the generic and personalized \gls{fl} objectives, and follow up by outlining iterative parameter alignments unique approach.  

%by defining  algorithm, $\mathtt{FEDAVG}$, as well as the objective personalized \gls{fl} algorithms.  , followed by defining our objective and framework in Section 2.2.  

\textbf{Background.}
In standard \gls{fl} there are $N$ clients in a federation, where each client $i$ has a local dataset $\mathcal{D}_i $.  
The goal is to solve a common objective over  a universal dataset $\mathcal{D} = \cup_{i \in [N]}$ by aggregating each local model into a global model.  The system iterates between local training on each client and global aggregation at the server.  \gls{fa}, the original \gls{fl} algorithm \cite{mcmahan2017communication}, involves a weighted averaging of client parameters at the server: 
%\vspace{-.1em}
\begin{equation}
    \textbf{Local :}\quad \theta_i=\argmin_{\theta\in\mathbb{R}}\mathcal{L}_i(\mathcal{D}_i; \theta)
    \text{, initialized with } \theta 
\end{equation}
%\vspace{-.1em}
\begin{equation}
 \textbf{Global:}\quad 
  \theta=  \sum_{i=1}^{N} \frac{|\mathcal{D}_i|}{|\mathcal{D}|} \theta_i
\end{equation}

where $\theta_i$ is the local model's parameters, $\theta$ is the global model's parameters, $\mathcal{L}_i(\theta) = \mathbb{E}_{(x,y)\sim\mathcal{D}_i} \bigl[  \ell_i(f(x),y;\theta)\bigr]$ is the local empirical loss of model $i$ on dataset $\mathcal{D}_i $, and $x$ and $y$ are the samples and labels in $\mathcal{D}_i$.   

\textbf{Iterative Parameter Alignment.}
To begin, we consider a set of $N$ peers (rather than clients)  where peer $i$ has access to local dataset $\mathcal{D}_i $. Similar to standard \gls{fl}, our goal is to solve an objective over universal dataset $\mathcal{D}$ for each peer model $f(\theta_i)$.
%as well as the full set of peer parameters $\theta^*$.   %We minimize the local empirical loss $\mathcal{L}^\mathnormal{E}_\mathnormal{i}$ for parameters $\theta_i$: 
To do this, each peer solves both an empirical learning objective, denoted $\mathcal{L}_\mathnormal{i}$, as well as an alignment objective $\mathcal{A}_\mathnormal{i}$, which together minimize the set of \textit{all} peer parameters $\theta^*$, where $\theta^*=\{\theta_1,...\theta_N \}$: 

%\vspace{-.5em}
\begin{equation*}\label{e2}
 \argmin_{\theta^*\in\mathbb{R^d}} \bigl[ \mathcal{L}_i(\mathcal{D}_i;\theta_i) + \mathcal{A}_i(\theta^*) \bigr] 
\end{equation*}
%\vspace{-.3em}

where $\mathcal{L}_i(\theta_i) = \mathbb{E}_{(x,y)\sim\mathcal{D}_i} \bigl[  \ell_i(f(x),y;\theta_i)\bigr]$ and $x$ and $y$ are samples from $\mathcal{D}_i$.  For experiments in this work, we set $\ell$ to be a cross-entropy loss for image classification problems.  Importantly, $\mathcal{D}_i$ is only seen by peer model $f(\theta_i)$, from which the empirical loss is calculated. Moreover, peers are not able to share data with each other, only model parameters. This is similar to parameter sharing among the client and server in standard \gls{fl}. We can apply differential privacy to the parameter sharing similar to previous work \cite{wei2020federated}, albeit in a decentralized (rather than centralized) topology.

Key to the global convergence of a peer model is the alignment of parameters during training.  Specifically, model $i$ holds parameters $\theta^* $ locally, and during each minibatch updates $\theta_i$ by minimizing the distance between $\theta_i$ and \textit{each} $\theta_n$.  For a single weight matrix or bias for model $i$ we denote this as $\mathnormal{a}_i$:  
%\vspace{-.2em}
\begin{equation}
    \mathnormal{a}_i(\theta^*)=\sum_{n=1}^N ||\theta_i-\theta_n||_p,  \ \text{where} \ i \neq n
\end{equation}
%\vspace{-.3em}

where $p$ is the $L_1$ or $L_2$ distance.  In other words, $\mathnormal{a}_i$ is the sum of distances in $\theta^*$ between $\theta_i$ and $\theta_n$. 
 Generalizing parameter alignment across the weights and biases of each layer $1,l,...L$ of a neural network we achieve our alignment objective for model $i$:
%\vspace{-.3em}
\begin{equation}
    \mathcal{A}_i(\theta^*)= \ \lambda \sum_{l=1}^L  \mathnormal{a}_i(\theta^*)
\end{equation}\label{align_eq}
%\vspace{-.1em}
where $\lambda$ is a global scale factor on the weight alignment objective.  We set $\lambda$ to 1 in this work.  
\gls{ipa} leads to a minimization of the global loss in individual models who have never seen the global dataset.  In other words, when solving for the alignment objective in Equation \ref{align_eq},
 we show that a peer model with access to the full parameter set $\theta^*$ iteratively converges to an objective solved over the global dataset $\mathcal{D}$: $ \argmin_{\theta_i} \mathcal{L}_i(\mathcal{D}_i;\theta_i) \rightarrow \argmin_{\theta} \mathcal{L}(\mathcal{D};\theta)$.  
 Compared to standard \gls{fl}, the \gls{ipa} algorithm only updates parameters on peer devices in a decentralized and synchronous architecture.   Further, the method relies on independent (i.e. never aggregated) peer models.  In the next section, we highlight the benefits of the approach in various settings.  
 
% Practically, this has several distinctions from standard \gls{fl}: 1)  Parameters are only updated on peer devices.   and 2) Local updates are regularized by optimal solutions of peer models. 

%\begin{equation*}
   % \forall\quad1 \leq i \leq N
% \argmin_{\theta_i} \mathcal{L}_i(\mathcal{D}_i;\theta_i) \rightarrow \argmin_{\theta} \mathcal{L}(\mathcal{D};\theta) %\ \forall \ i \in N 
%\end{equation*}

% this is not saying a whole lot in the paragraph.  
%where global loss $\mathcal{L}(\theta)=\mathbb{E}_{(X,Y)\in\mathcal{D}} \  \mathnormal{l}(f(X),Y;\theta)$, and $\theta^*$ represents the full parameter set $\theta_{i \in N}$.  $X$ and $Y$ are the full sample and label set in dataset $\mathcal{D}$. 
%For example, given models $f_i$ and $f_j$ which have learned some objective using subset datasets $\mathcal{D}_1$, $\mathcal{D}_2$ in $\mathcal{D}$, we can \textit{merge} such models in parameter space such that $\mathcal{L}_i(\theta_i)=\mathbb{E}_{(X,Y)\in\mathcal{D}} \  \mathnormal{l}(f_i(X),Y;\theta_i)$ and 
 %$\mathcal{L}_j(\theta_j)=\mathbb{E}_{(X,Y)\in\mathcal{D}} \  \mathnormal{l}(f_j(X),Y;\theta_j)$.  
%This scenario works for both IID and non-IID data partitions.  

      \begin{algorithm}[H]
        \caption{Parameter Alignment, One Iteration.  }
        \begin{algorithmic}
         \State \textbf{Input:}
\Indent
\State $N$ peers
\State Peer\textsubscript{$i$} has:
dataset $\mathcal{D}_i$, model $f_i$ with weights $\theta_i$
     
     \State $\theta^*$  is all peer parameters: $\{\theta_i,...\theta_N\}$
    %\EndIndent
\EndIndent
\State \textbf{Output:}
\Indent
\State Models $f_1(\theta_1),f_2(\theta_2),...f_N(\theta_N)$
\EndIndent
\State
%\State \textbf{Execute:}  
  %  \Indent
    \State Each peer initializes $\theta_i$, sends to \texttt{peer}\textsubscript{1}
    
\ForEach { \texttt{peer}\textsubscript{$i$} $ \in N $}  
    \ForEach {batch $b \in \mathcal{D}_i$}  
   \State $\mathcal{L}_i=\ell(f_i(b;\theta_i))+\Call{\textbf{ParamAlign}}{f_i, \theta^*}$

    \State $\theta_i \gets \theta_i- \triangledown\mathcal{L}_i$
   % \State $\theta_i \gets \theta_i- \triangledown (\ell(f_i(b;\theta_i))+\Call{\textbf{WeightAlign}}{f_i, \theta^*})$
\EndFor
\State \texttt{Transfer $\theta^*$ to peer\textsubscript{$i+1$} }
\EndFor
 %   \EndIndent
\State 
\Function{\textbf{ParamAlign}}{$f_i$, $\theta^*$ }: 
\Indent
\State $\mathcal{R}_i \gets 0$
 \ForEach {layer $ \in f_i$} 
 \ForEach {$\theta_j \in \theta^*, j \neq i $}  
  \State $\mathcal{R}_i \gets \mathcal{R}_i + |\theta_i - \theta_j|_p$ 
\EndFor
\EndFor
\State \Return $\mathcal{R}_i$
\EndIndent
\EndFunction

        \end{algorithmic}
      \end{algorithm}

\section{Experiments}\label{experiments}

We begin by evaluating Iterative Parameter Alignment against existing methods in federated learning, including experiments merging peer models trained on segregated classes. Next, we quantify the difference between peer models, showing that each peer produces a distinct model in both parameter space and during inference. Finally, we highlight the ability for \gls{ipa} to produce fair models (at epoch $t$), converging thereafter to globally optimized solutions. %Our code is publicly available at \url{https://github.com/mattgorb/iterative_weight_alignment}.

\begin{figure*}
\centering
   \includegraphics[width=1\linewidth]{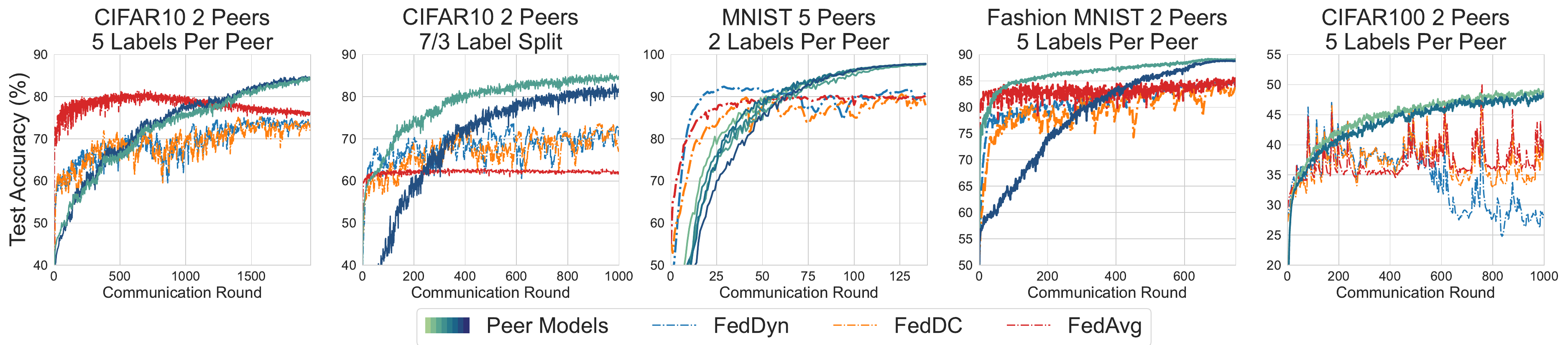}
   %\caption{}
   \label{fig:disjoint} 
\caption{\textbf{Aligning Peer Models Trained on Disjoint Classes:}  We find that existing federated learning approaches such as \gls{fa} struggle when trying to merge models trained with divergent (rather than heterogeneous) data partitions. We show how \gls{ipa} achieves stable training compared to existing approaches, with \gls{ipa} eventually converging to baseline accuracy compared to other methods which create global models with unstable performance. %For $\mathtt{FedDyn}$ and $\mathtt{FedDC}$ we apply smoothing to the test accuracy as a result of the instability of the global model test accuracy. 
}\label{disjoint}
% \vspace{-0.5em}
\end{figure*}

%  \vspace{-1.2em}

\subsection{Domain Divergent Silos}
%\vspace{-0.2em}
Unique to this work we experiment with merging peer models that have completely segregated classes. For example, Peer$_1$ may only have images of dogs while Peer$_2$  has only images of cats.  Such scenarios are important in the real-world such as those involving GDPR where an entire demographic segment is isolated, or cross-industry learning where the domains of individual peers are disjoint.

The scenario also highlights the distinction between \gls{ipa} and personalized \gls{fl} \cite{chen2021bridging, chen2023the}.  In the above example, personalized \gls{fl} would aid Peer$_1$ to better generalize to its own domain (dogs) by utilizing Peer$_2$ data. However, Peer$_1$ may not gain much value from Peer$_2$'s information about cats.  We further distinguish \gls{ipa} from personalized \gls{fl} in Section \ref{personalized_related}. 

\gls{ipa}, in contrast, can successfully merge two or more seemingly independent domains. Figure \ref{fig:arc} (center) shows how three peers trained with different CIFAR-10 labels can be iteratively aligned and each converge to the accuracy of a baseline model trained on all data. 

We compose our experiments with simple class splits, such as a two-peer class split where one peer has all training data labeled 0 to 4 and the second peer has training data labeled 5 to 9 (in a dataset with 10 classes).  We also consider imbalanced splits such as peers with an unequal number of classes.

%Our experiments include a two peer class split and a five peer class split.  For the two peer class split, one peer has all training data labeled 0 to 4 and the second peer has training data with labeled 5 to 9 (in a dataset with 10 classes).  For the five peer class split, we split training data using its associated labels into five distinct training sets, one for each peer.  We present results using the CIFAR10 and CIFAR100 datasets.  

\textbf{Results.} 
Figure \ref{disjoint} highlights the convergence of peer models trained using the \gls{ipa} algorithm on disjoint classes.  We find that compared to \gls{fa}, $\mathtt{FedDyn}$, and $\mathtt{FedDC}$, \gls{ipa} achieves stable training.  
Moreover, under both balanced splits (each peer has the same number of labels) and imbalanced data splits, \gls{ipa} converges consistently to baseline accuracy.  Existing algorithms such as  $\mathtt{FedDyn}$ and $\mathtt{FedDC}$ have very unstable training; their global model test accuracy curves were smoothed in Figure \ref{disjoint} for visualization purposes.  \gls{fa} had more stable training, however, its naive parameter averaging technique did not converge to baseline accuracy.  Rather, its performance flattened out after a few communication rounds.  

We hypothesize that existing \gls{fl} algorithms are unstable in the segregated class scenario because the gradient updates of local models are entirely disassociated from each other as a result of the domain discrepancy. 
Existing work has shown that clients with heterogeneous data partitions have inconsistent optimization directions \cite{khaled2020tighter, karimireddy2020scaffold}, which cause drifts in the local models away from a global solution.  We tried over half a dozen different configurations for existing algorithms, including different seeds, reduced learning rate, and a smaller number of local epochs. %We include additional results in the Appendix.  

%\vspace{-1em}
\label{segregate}

%\vspace{-.4em}
\subsection{Comparison to Existing Approaches}
%\vspace{-.2em}
Our second empirical study compares the convergence rate of Iterative Parameter Alignment against existing \gls{fl} algorithms.  McMahan et al. \cite{mcmahan2017communication} noted the slow convergence of \gls{fa} when clients had heterogeneous data partitions.  
Since the initial research, much effort has been put into improving this convergence rate,
which is measured by the \textit{number of communication rounds} between the clients and the server until the global model reaches some target accuracy on the test set.  We test \gls{ipa} in a similar fashion, where one communication round equals each peer performing their allotted training. 

%\begin{wrapfigure}[13]{r}{0.5\textwidth}
\begin{table}
\setlength{\tabcolsep}{0.4em}
\scalebox{.98}{
  \begin{tabular}{lccccccc}
\toprule
\toprule
    Dataset &  \thead{Target \\ Acc. (\%)} & FedAvg&FedProx&Scaffold& FedDyn& FedDC & \gls{ipa}\\
\toprule
\toprule

    \multicolumn{8}{c}{IID, 20 Peers, $p=2$}\\
        \midrule
													
MNIST	&	98	&	49	&	46	&	50	&	20	&	33	&	3	\\
Fashion	&	89	&	148	&	151	&	165	&	35	&	100	&	14	\\
CIFAR-10	&	85	&	42	&	46	&	31	&	20	&	20	&	15	\\
CIFAR-100	&	50	&	82	&	84	&	45	&	60	&	43	&	30	\\
\toprule
    \multicolumn{8}{c}{Dir. ($\alpha=0.6$), 20 Peers, $p=1$}\\											\midrule
MNIST	&	98	&	147	&	140	&	52	&	20	&	35	&	28	\\
Fashion	&	87	&	60	&	67	&	62	&	15 &	40	&	60	\\
CIFAR-10	&	85	&	64	&	65	&	44	&	22	&	24	&	44	\\
CIFAR-100	&	50	&	105	&	105	&	56	&	61	&	55	&	97	\\
\toprule
    \multicolumn{8}{c}{Dir. ($\alpha=0.3$), 20 Peers, $p=1$}\\
    \midrule
MNIST	&	98	&	139	&	199	&	57	&	45	&	39	&	70	\\
Fashion	&	87	&	98	&	93	&	92	&	25	&	50	&	90	\\
CIFAR-10	&	85	&	133	&	144	&	58	&	28	&	29	&	95	\\
CIFAR-100	&	50	&	111	&	110	&	64	&	74	&	55	&	103	\\
                    \midrule
   % \multicolumn{8}{r}{\scriptsize{
  %  \textbf{1^{st}}, \textcolor{red}{2^{nd}}, \textcolor{blue}{3^{rd}}
  %  }
  %  }
    
   %\\

  \end{tabular}
  }
  \makeatletter\def\@captype{table}\makeatother
  \caption{\textbf{Communication rounds required to achieve target accuracy: } We compare the number of communication rounds required for \gls{ipa} and other state-of-the-art \gls{fl} algorithms to reach a target accuracy.  \gls{ipa} converges quickly on IID data, with competitive results on heterogeneous splits. \gls{ipa} does not achieve state-of-the-art performance in heterogeneous experiments, however, communication is less of a constraint in cross-silo settings. 
 }\label{statistics}

\end{table}
%\end{wrapfigure}

\textbf{Experimental Setup.} We construct our experiments 
from a set of scenarios with homogeneous and heterogeneous data partitions consistent with previous research.  In heterogeneous settings, our label ratios follow the  Dirichlet distribution with $\alpha=0.3$ and $\alpha=0.6$, similar to previous works.  Lower $\alpha$ indicates a higher data heterogeneity.  
We compare Iterative Parameter Alignment to the standard \gls{fl} algorithm \texttt{FedAvg} \cite{mcmahan2017communication} as well as state-of-the-art approaches \texttt{FedProx} \cite{li2020federated}, \texttt{Scaffold} \cite{karimireddy2020scaffold}, \texttt{FedDyn} \cite{acar2021federated}, and \texttt{FedDC} \cite{gao2022feddc}.  The original hyperparameters are used for each algorithm. We compare algorithms using MNIST, FashionMNIST, CIFAR-10, and CIFAR-100 datasets. We use the same architecture as previous works for the MNIST and FashionMNIST datasets; for the CIFAR-10 and CIFAR-100 datasets, we use a larger CNN model which includes four convolutional layers followed by three linear layers.  We consider one round of communication as each client training the model and sending it back to the server for aggregation (100\% client participation).  For \gls{ipa}, we report the number of communication rounds it takes for the first peer to reach a target accuracy. 

Unique to Iterative Parameter Alignment, we report the convergence rates of peer models with \textit{different initializations}, i.e. each peer model is initialized from a different random seed.  In the original \gls{fl} work, the authors highlighted the success of naive parameter averaging when models had the same initial weights.  Averaging did not perform as well when the models were initialized differently.  This phenomenon was also reported in model merging literature \cite{matena2022merging}, where the authors required models trained from the same initial weights.  
Research has suggested permutation invariance of neural networks as a driving force for this observation, i.e. a neural network has many variants which differ only in the ordering of its parameters \cite{wang2019matcha}.  %We conjecture that \gls{ipa} moves weights into the same permutations early in alignment.  

\begin{figure*}
\begin{minipage}[s]{.6\textwidth}
    \renewcommand{\arraystretch}{1.3}%
       \scalebox{1.1}{
  \begin{tabular}{ccccc}
    &\multicolumn{2}{c}{FashionMNIST}&\multicolumn{2}{c}{CIFAR-10}\\
    \toprule
          \toprule
  Distance   & Dir(0.3) & IID & Dir(0.3) & IID \\
    \midrule
          \toprule
 $|| \theta_i-\theta_j||_1$& 196.5 & 1352.6 &$1.8\times10^3$&3.3$\times10^4$ \\   
   $|| \theta_i-\theta_j||_2$& 0.7 & 4.9 & 2.0 & 35.9  \\   
  $\mathcal{H}(f_i, f_j)$ &1,990& 650 & 2,504 & 1,043  \\   
    $f_i \wedge f_j$&7,358 &8,603& 6,871& 8,214 \\  
    $ \textoverline{$f_i$} \wedge \textoverline{$f_j$} $& 947& 837 & 1,094 & 947  \\  
    \bottomrule
  \end{tabular}
  }
\end{minipage}%\hfill
\begin{minipage}[s]{.4\textwidth}       
\scalebox{0.9}{
  \centering
  \includegraphics[width=1\linewidth]{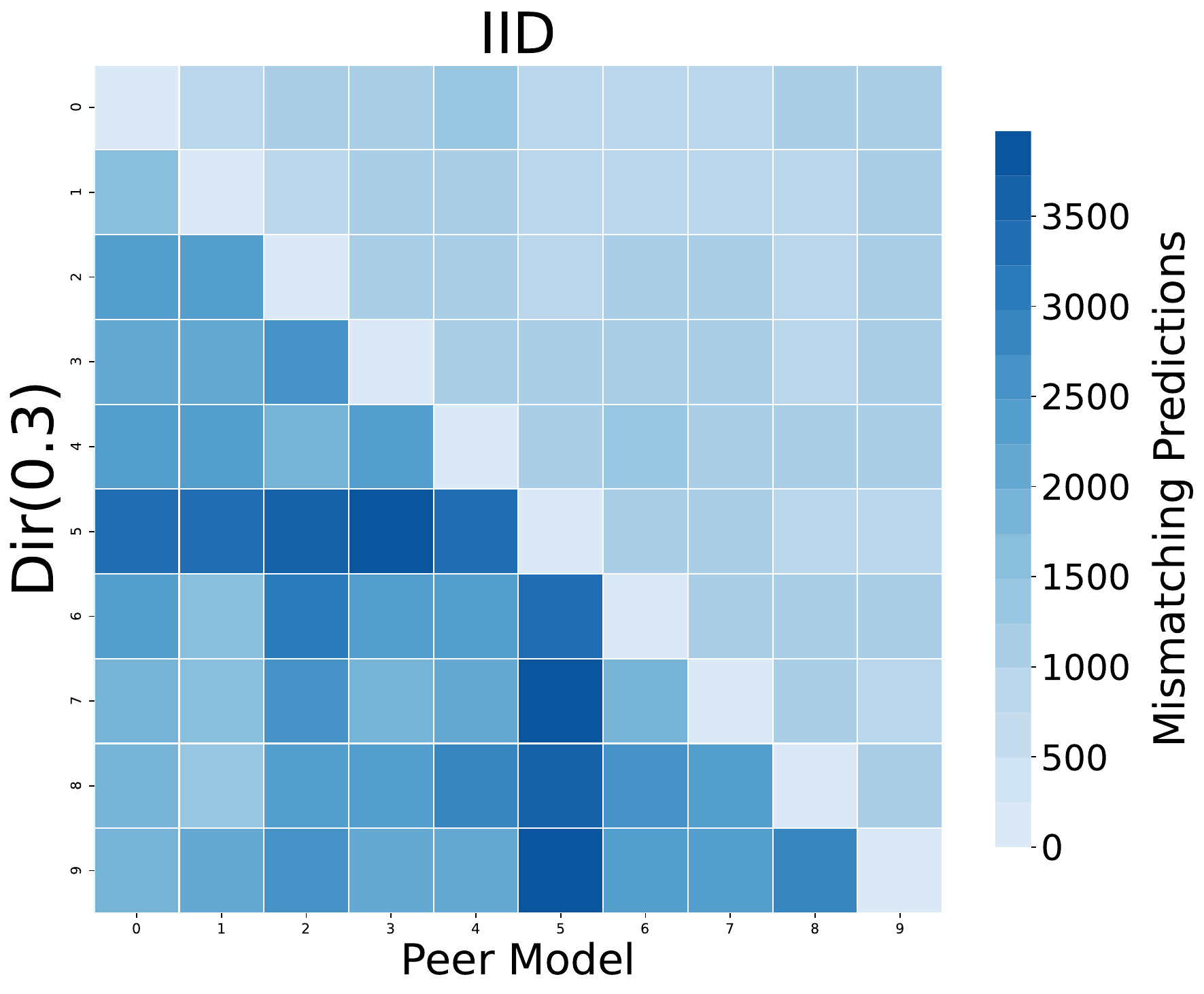}
  }
\end{minipage}
\caption{\textbf{Comparing Peer Models:}  We measure the distance between peer models across a variety of metrics.  Each experiment contains ten peers and is aggregated across three runs, with the mean presented for each.  \textbf{Left:}  Measuring the distance between models across parameters (first two rows) and model predictions (the last three rows). The last three rows denote the Hamming distance between predictions, mutual correct predictions, and mutual incorrect predictions on the test set.  Test set size for both datasets is 10k. \textbf{Right:} A similarity matrix of Hamming distances between peer model predictions for: 1) heterogeneous data partition (bottom triangle) and 2) homogeneous (IID) data partition (top triangle). The distances represent the number of mismatching predictions in the test set for each model.  For reference, the lowest (averaged) Hamming distance between models in the IID setting is 880, with a test set size of 10k.  }\label{differences}
\end{figure*}

\textbf{Results.}
Table \ref{statistics} highlights the results of \gls{ipa} against five state-of-the-art methods.  Unsurprisingly, under IID settings \gls{ipa} converges quickly towards the target accuracy on all four datasets.  While the algorithm only feeds dataset $\mathcal{D}_i$ to $f(\theta_i)$, it  has $20 \times\bar{\theta}$ parameters, optimizing $19\times\bar{\theta}$ parameters using alignment and the final $\theta_i$ using alignment plus empirical loss.  As a result, the balanced, overparameterized networks converge quickly despite only having access to a fraction of the training samples. Compared to existing approaches, \gls{ipa} achieves state-of-the-art performance.  

Under increasingly heterogeneous settings (from top to bottom) we observe a longer convergence rate for \gls{ipa} compared to other algorithms.  \gls{ipa} remains competitive for MNIST and FashionMNIST, however, has a slightly longer convergence rate for CIFAR-10 at Dirichlet ($a=0.3$) as well as CIFAR-100. We argue that convergence rate is less of a concern in cross-silo settings since large companies likely have adequate computation.  Moreover, it achieves better accuracy than baseline models  FedAvg and FedProx. %Further, data heterogeneity creates additional questions in collaborative learning which we discuss in Section \ref{fairness}. 

\subsection{Peer Model Comparison}\label{comparison}

%https://ieeexplore.ieee.org/abstract/document/963772?casa_token=lLIKrdVBkc8AAAAA:6NBe01exSU96xSufqy3aB7Z-tFuZkqTVpClTMis-G3hgXmsr2NhyT3flwi86bSAp9HrpAkaQnA

%

%In this section we assess the quantitative differences between peer models generated by \gls{ipa}. 
%Our approach creates unique models for each peer, the alignment of parameters during train time implies similarity. 
%Despite the potential similarity of peer models as a result  
%Since our approach creates unique models for each peer, 
%Existing literature has found that neural networks are known to be sensitive to small perturbations in their parameters \cite{weng2020towards} which cause drastic changes in model inference and generalization.  The  phenomena has been used for injecting adversarial attacks on models \cite{zhao_pu,8203770,weng2020towards}, evaluating the generalization gap of model minima \cite{neyshabur2017exploring, keskar2016large}, and assessing the effects of model quantization \cite{hubara2017quantized}. 

We look at the quantitative differences between peer models across a variety of metrics to assess whether \gls{ipa} creates \textit{sufficiently unique} models.  Existing literature has found that neural networks are known to be sensitive to small changes in their parameters \cite{weng2020towards}, causing drastic changes in model inference and generalization.  There is a rich area of research examining this phenomenon for injecting adversarial attacks \cite{zhao_pu,8203770,weng2020towards}, evaluating the generalization gap of model minima \cite{neyshabur2017exploring, keskar2016large}, and assessing the effects of model quantization \cite{hubara2017quantized}. As a result, even the smallest differences in the weights of peer models can create unique results.

\textbf{Experiments.} To quantify the difference between two peer neural networks we compare both the network parameters as well as the predictions.  We measure the distance between two models' parameters as $||\theta_i- \theta_j||_p$, where $p=\{1,2\}$. To measure the difference between model predictions, we compute the Hamming distance between models' outputs on the test set, which we denote $\mathcal{H}(f_i, f_j)$.   We also present a count of when both models' predictions are correct (denoted $f_i \wedge f_j$), as well as both incorrect ($ \textoverline{$f_i$} \wedge \textoverline{$f_j$} $). 
%We normalize all predictions by dividing them by the size of the test set.  

We test heterogeneous (Dirichlet with $\alpha=0.3$) and homogeneous scenarios with both the FashionMNIST and CIFAR10 datasets.  All experiments use ten peers and are averaged over three runs.  We choose a lower number of peers compared to previous experiments in order to magnify potential similarities between models. Heterogeneous experiments are trained for 200 epochs and homogeneous experiments are trained for 50 epochs. 
The FashionMNIST experiment on homogeneous data had a test accuracy of 88.9\%$\pm0.24$, and the heterogeneous scenario 82.5\%$\pm3.42$. The CIFAR10 experiment on homogeneous (IID) data had a mean test accuracy of 86.4\%$\pm0.44$, and the heterogeneous scenario  79.5\%$\pm4.12$.

\textbf{Results. } Table \ref{differences} highlights the differences between peer models across four experiments.  The first two rows indicate a dissimilarity between peer model parameters across $L_1$ distance, with a smaller discrepancy  when measured with $L_2$ distance.  We hypothesized that IID data experiments would have closer parameters, however, the heterogeneous experiments yielded smaller values.  We speculate this is because we train heterogeneous data for 200 epochs compared to just 50 epochs for IID data.  

The bottom three rows measure the difference in test inference between peer models, with both datasets having a test set size of 10k.  The smallest Hamming distance was between IID models, with 650 for FashionMNIST and 1,043 for CIFAR-10.  We argue that these values indicate a significant difference from each other since IID models achieve 88.9\% and 86.4\% accuracy on the test set.  Finally, we note that the standard error was negligible across all experiments.

%\url{https://cs.stackexchange.com/questions/74488/measuring-difference-between-two-sets-of-neural-network-weights}

%Visual this with dissimilarity matrix 1. during training 2. training completion.  

%Experiments: CIFAR-10/10 clients/also use baseline method with all training data

%Key to Iterative Weight Alignment is the creation of \textit{unique} models for each peer.  In this section we measure the individualized properties of each model produced using the algorithm.  

%To begin, we consider two experimental setups, each consisting of ten peers: the first using the CIFAR-10 dataset with a Dirichlet 

\subsection{Fairness through Early Stopping } \label{fairness_section}

\begin{table}
\large
\begin{center}

    \renewcommand{\arraystretch}{.98}%
   \scalebox{0.98}{
  \begin{tabular}{ccccc}
      \toprule
          \toprule
    &\multicolumn{2}{c}{MNIST}&\multicolumn{2}{c}{CIFAR-10}\\
        &\multicolumn{2}{c}{ 20 Peers}&\multicolumn{2}{c}{10 Peers}\\
    \toprule
          \toprule
  Algorithm   & CLA& POW& CLA&POW \\
    \midrule
          \toprule

q-FFL \cite{li2019fair}	&	38.7	&	48.07	&	51.33	&	94.06	\\
CFFL \cite{lyu2020collaborative}	&	94.7	&	85.71	&	72.55	&	81.31	\\
ECI	\cite{song2019profit}&	99.41	&	95.21	&	79.5	&	99.55	\\
CGSV ($\beta$=1) \cite{xu2021gradient}	&	96.39	&	97.23	&	98.78	&	99.89	\\
CGSV ($\beta$=2)	\cite{xu2021gradient}&	91.33	&	94.32	&	88.78	&	93.39	\\
\textbf{\gls{ipa}} (Ours)&96.44&95.98 &95.86&	92.22	\\

  \end{tabular}
  }
  \caption{\textbf{Fairness of \gls{ipa} compared to existing approaches: } We compare the fairness of various collaborative learning approaches against \gls{ipa} by measuring the \textit{correlation} of each clients model performance compared to its standalone models' performance. 
 Correlation is scaled between -100 and 100.  }\label{fairness1}
\end{center}
\end{table}

In cross-silo settings organizations may be competing against each other, hence the contribution of participants becomes a critical measure.  Designing proper incentive mechanisms and rewards for participation can encourage peers to join a federation. 
Previous work has proposed fairness schemes using methods such as contract theory \cite{8832210,kang2019incentive}, monetary payouts \cite{yu2020sustainable}, game-theoretic approaches \cite{donahue2021optimality, blum2021one}, the Shapely value \cite{xu2021gradient,10.1145/3501811}, and reputation lists \cite{lyu2020collaborative}. Most of these methods produce variations of the single global model, i.e. models for each client whose performance is commensurate to its data contribution.  

 \gls{ipa} takes a different approach: Figure \ref{fig:arc} highlights the variable convergence rates of peer models with heterogeneous data partitions.  We find that the convergence of a peer model  trained with \gls{ipa} is a function of the peers' standalone model performance. We enable fairness in the \gls{ipa} algorithm through the \textit{early stopping} of training at some iteration $t<T$, where $T$ is the number of iterations it takes for \textit{all} peer models to converge to some target accuracy. %Typically, early stopping is performed in neural network training to avoid overfitting; we instead use it to provide fairness to the federation so that each member receives a model whose performance is similar to their contribution.  

%Iterative Parameter Alignment produces an \textit{in-built} incentive measurement as a function of peer data heterogeneity as well as data quantity.  
\begin{comment}

\begin{wraptable}{l}{0.4\textwidth}
\caption{A wrapped table going nicely inside the text.}\label{wrap-tab:1}
\begin{tabular}{ccc}\\ 
\toprule
\toprule
Algorithm &  POW  & CLA \\
\midrule
\toprule
    \multicolumn{3}{c}{MNIST, 20 Peers }\\
    \midrule
    MNIST, 20 Peers&1&1\\
    MNIST, 10 Peers&1&2\\
    CIFAR10, 10 Peers&3&2\\

\bottomrule

\end{tabular}
\end{wraptable} 
\end{comment}

To test our approach, we conduct experiments designed from benchmarks in previous works.  We measure fairness using a scaled Pearson's coefficient: $100\times\rho(\varphi, \xi)\in [-100, 100]$ \cite{lyu2020towards, xu2021gradient, lyu2020collaborative}.  Specifically, we measure the correlation between the test set accuracy of the set of standalone models ($\varphi$) compared to the test set accuracy of the set of models generated by \gls{ipa}  ($\xi$) .   \textit{The intuition is that peers should have a federated model with similar capabilities to their standalone model relative to others peers}. 

Our first experiments involve comparing our method with the benchmarks of Xu et al. \cite{xu2021gradient} since their approach provides theoretically guaranteed fairness metrics.  Additionally we compare q-FFL \cite{li2019fair}, CFFL \cite{lyu2020collaborative}, and ECI \cite{song2019profit}.  Experiments use the CIFAR10 and MNIST datasets and apply class-imbalanced (CLA) and size-imbalanced (POW) data partitioning each using 600 samples per peer.  For \gls{ipa}, we run each model in a random topology with one local training epoch per peer since there is a limited amount of training data.  For CIFAR10, we average peer model performances on the test set for epochs 5-15, while in MNIST we average peer model performances in epoch 1-5.  
Results in Table \ref{fairness1} show that \gls{ipa} has a correlation above 86 for each of the four tests, with three of the results above 95.  These metrics are on average stronger than each prior method except for CGSV at $\beta=1$.  %We also visualize these results in Figure \ref{fairness1}.    

\begin{figure}
\begin{minipage}[s]{.235\textwidth}
  \centering
  \includegraphics[width=1\linewidth]{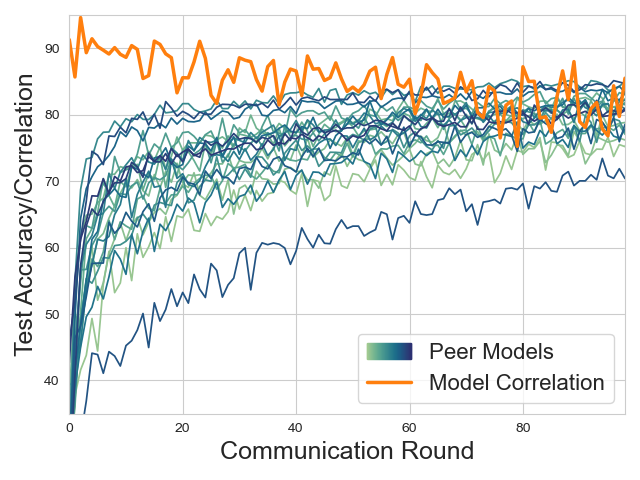}
\end{minipage}%
\begin{minipage}[s]{.23\textwidth}
  \centering
  \includegraphics[width=1\linewidth]{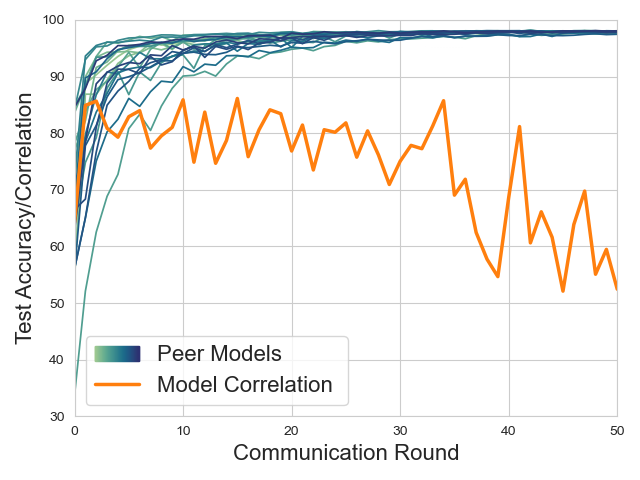}
\end{minipage}%
\caption{\textbf{Fairness across iterations: } We show that the \gls{ipa} algorithm creates fair models earlier in training before the global convergence of peer models. 
 CIFAR-10 (Left) and MNIST (Right) performance across models and communication rounds, overlaid with peer models' correlation with their standalone performances (orange).  We find that early in training, peer model performances are correlated with the performance of their standalone models relative to other peers.  As training proceeds and the peer models globally converge, model fairness decreases, as can be seen in the MNIST figure.   }\label{fairness2}
\end{figure}

Next we experiment with a more robust and realistic data partitioning by using the full CIFAR-10 and MNIST datasets, 20 peers, and a Dirichlet split with $\alpha=0.25$.  We run each experiment four times in a random topology and test the correlation between \gls{ipa} and standalone model performance.  In these experiments, we find that the \textit{test loss} (rather than the test accuracy) is a stronger metric for correlation.  For CIFAR-10, we average the test loss between communication rounds 50 and 100 to gain a thorough picture of the correlation, and to counter the variance of individual communication rounds.  For MNIST, we average communication rounds 5 to 25.  Overall our CIFAR-10 experiments have a correlation of 86.3 $\pm2.2$, while our MNIST experiments have a correlation of 80.5 $\pm3.4$. Figure \ref{fairness2} depicts the test accuracy of peer models across communication rounds overlaid with the correlation (orange) of the group of peer models compared to the group of standalone models.

Finally, we would like to note that \gls{ipa} offers a distinct advantage in homogeneous settings: in existing fairness approaches, peer models will be more or less identical as a result of being derived from the same global model. \gls{ipa}, however, will produce a unique solution for each peer in the homogeneous setting. \label{fairness}

\vspace{-0.2em}
\section{Additional Analysis}
\vspace{-0.2em}

\textbf{Simulating Differential Privacy} To simulate differential privacy we run two experiments which test the effect of adding noise to peer model parameters.  Our motivation is to test whether peer models still converge to a high test set accuracy when differential privacy is applied to each peer. Specifically, when one peer is finished with a training iteration, we add a small amount of noise to their parameters ($\theta_i$) prior to sharing with others.  We add random noise with $\mu=0$ and $\sigma=0.0005$. 

Our first experiment uses the CIFAR-10 dataset with two silos, each with half of the labels.  Both models converge to 85\% test accuracy after roughly 4,000 rounds.  Our second experiment uses CIFAR-10 with 20 peers using a Dirichlet data split with $\alpha=0.6$.  The first silo converges to 85\% test accuracy after 197 rounds.  While both of these are significantly longer than \gls{ipa} without differential privacy, differential privacy provides security guarantees for each silo.  %Moreover, \gls{ipa} is desireable  communication and computat

\begin{figure}[H]
\begin{minipage}[s]{.235\textwidth}
  \centering
  \includegraphics[width=1\linewidth]{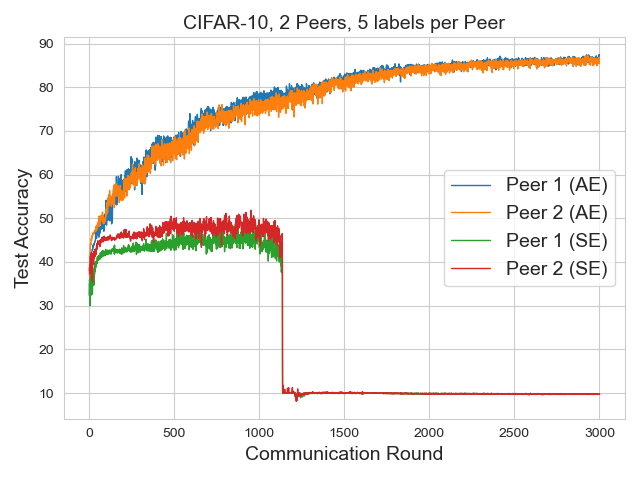}
\end{minipage}%
\begin{minipage}[s]{.23\textwidth}
  \centering
  \includegraphics[width=1\linewidth]{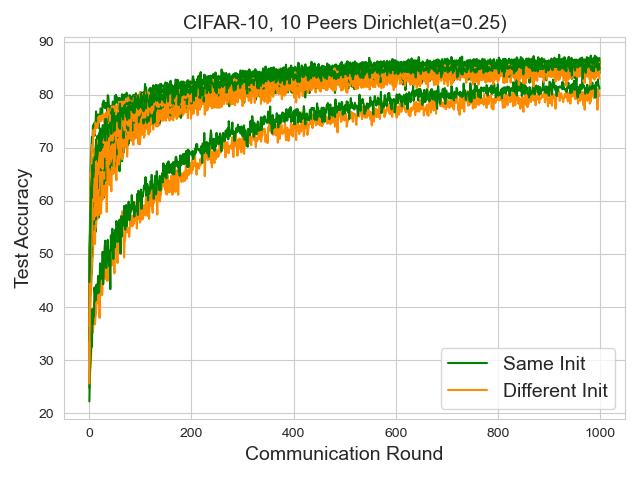}
\end{minipage}%
\caption{\textbf{Left:} We show the instability of squared error alignment compared to absolute error in a CIFAR-10 experiment with two peers with 5 labels each. \textbf{Right:} We discover similar convergence rates of peers when models have the same initialization (green) and different initializations (orange).}\label{fig:heterogeneous_compare}
\end{figure}

\textbf{$L_1$ versus $L_2$ Parameter Alignment.}  In Figure \ref{fig:heterogeneous_compare} (left) we show that split label experiments exhibit instability when using squared error ($L_2$) alignment, while absolute error ($L_1$) alignment achieves smooth convergence.  We observed similar results in all experiments using heterogeneous and disjoint data partitions.  

%Figure \ref{fig:homo_compare} shows the convergence of both absolute error and squared error under homogeneous settings for the CIFAR-10 dataset with ten peers.  Both absolute error and squared error converge very quickly in this setting, with absolute error showing a slight advantage.

\textbf{Effect of Initialization Strategy.} We show how models are able to be aligned even when they have different initializations.  In Figure \ref{fig:heterogeneous_compare} (right), we show the convergence of a CIFAR-10 experiment with ten peers split with Dirichlet with $\alpha=0.25$.  
Both the green (same initialization) and orange (different initialization) converge at similar rates.

%\textbf{\gls{ipa} as an ensembling technique}

%\textbf{\gls{ipa} as an ensemble method}

\begin{comment}

\begin{figure}[h] %
    \centering
    \subfloat[\centering ]{{\includegraphics[width=6cm]{appendix/figs/ae_v_se_cifar10.png} }}%
    \qquad
    \subfloat[\centering ]{{\includegraphics[width=6cm]{appendix/figs/ae_v_se_mnist.png} }}%
    \caption{We show the instability of squared error alignment compared to absolute error with (a) CIFAR-10 with two peers with 5 labels each and (b) MNIST with 20 peers with a Dirichlet split with $\alpha=0.6$.  }%
    \label{fig:heterogeneous_compare}%
\end{figure}
\end{comment}\label{abl}

\section{Discussion}
\textbf{Limitations.}  \gls{ipa} is feasible specifically in cross-silo settings where peers have an adequate amount of computational capacity. It does not scale well to many peers as a result of requiring $N\times\bar{\theta}$ parameters during training unless all peers have large computational capacity. For example, \gls{ipa} worked well in our experimental settings with up to 20 peers on a single GPU where each model had 2-3 million parameters. 
We note that advances in neural network pruning and quantization may enable the method to scale well in the  
future \cite{frankle2018lottery, gorbett2023sparse,gorbett2023randomly}.  Additionally, methods in dataset and sample level measurement can enhance robustness and detection across difficult tasks \cite{gorbett2022wip,gorbett2022local,gorbett2023intrinsic, gorbett2022utilizing}. 

We note that \gls{ipa} works well under settings with \textit{reliable} peers. Standard \gls{fl} considers scenarios where peers go offline; we do not consider this scenario in this paper. However, if one peer drops out during the \gls{ipa} training process, their latest parameters will still be available for others to continue.  

There are additional settings we have not considered in this paper such as tasks other than image classification and vertically aligned \gls{fl} \cite{kairouz2021advances}.   

\textbf{Additional Security Considerations. } % in Real-World Settings.} %
Key to our approach is sharing model parameters across peers during the \gls{ipa} training process. While each peer produces an independent global model, each peer has access to others parameters during training.  This may lead to inadequate security and potential for misuse.  To counter this security flaw, we propose differential privacy on top of \gls{ipa}, which provides formalized privacy guarantees \cite{dwork2014algorithmic}.  Using this approach, each peer may add noise to their parameters before sharing with others.  Differential privacy is commonly applied to training data, however, it can also be applied to model training \cite{jagielski2020auditing}. We perform experiments on \gls{ipa} with differential privacy in Section \ref{abl}. 
 Differential privacy has been applied to the \gls{fl} pipeline \cite{mcmahan2017learning, geyer2017differentially, kairouz2021distributed, agarwal2021skellam} including in the cross-silo setting \cite{heikkila2020differentially} where additional considerations need to be made such as securing the privacy of sample-level (rather than client level) data \cite{liu2022on}. 

Homomorphic encryption \cite{paillier1999public} and garbled circuits \cite{c392ef4100b14c70b44c15b5b7a96531} are other protection techniques that enable peers to encrypt their models for enhanced protection; such techniques have been applied to \gls{fl} systems \cite{zhang2020batchcrypt, jiang2021flashe, 9130089}.  For example, homomorphic encryption allows clients to encrypt their model parameters before sending their updates to the server \cite{nvidia_homo}, effectively protecting their model against a potential malicious server.  Homomorphic encryption can be applied in a similar fashion in the \gls{ipa} algorithm, where peers send encrypted models to each other to hide the true values. %However, encrypted models may still be vulnerable

\textbf{Applications Beyond Federated Learning.}
\gls{ipa} may be of interest to other fields such as domain adaptation and transfer learning \cite{pan2010survey,yosinski2014transferable, wortsman2022robust, devlin2018bert, pruksachatkun2020intermediatetask}, model merging \cite{matena2022merging,ainsworth2022git}, model fusion \cite{pmlr-v119-hoang20b,pmlr-v139-lam21a,singh2020model}, ensembling \cite{wortsman2022model}, and other contexts with variable data distributions.
For example, %transfer learning enables fine-tuning a pre-trained model to enhance performance on some target domain. However, 
fine-tuning has been found to cause reduced robustness on source domain distribution shift benchmarks \cite{radford2021learning, pham2021combined}.  Wortsmann et al. proposed ensembling the pre-trained and fine-tuned models for increased performance on source domain robustness \cite{wortsman2022robust}.  Similar insights could potentially be gleaned from \gls{ipa}, where iteratively merging the parameters of segregated domains  provides enhanced performance.  Domain divergence is also an active area of research in negative transfer learning \cite{wang2019characterizing,zhang2022survey}, where source domain knowledge negatively effects a target domain's ability to learn. Exchanging and aligning models trained on divergent domains can enable opposing models to learn from each other, thereby enhancing generalization.  

%if the source and target domains diverge (known as negative transfer \cite{wang2019characterizing, zhang2022survey}), performance will degrade.  Moreover, 

%Domain adaption/transfer learning-negative transfer
%domains adaption: %https://openaccess.thecvf.com/content/CVPR2022/papers/Galstyan_Failure_Modes_of_Domain_Generalization_Algorithms_CVPR_2022_paper.pdf
%Since this works, we can 

\textbf{Conclusion} We propose a new method for iteratively aligning the parameters of peers models trained on independent data.  \gls{ipa} is favorable in segregated class settings, achieves state-of-the-art performance on homogeneous data partitions, and has competitive convergence under heterogeneous data partitions.  We assess our approach across novel and existing benchmarks and show that the method generates unique peer models that converge at a rate correlated to their standalone performance.

%Existing work in federated learning has largely concentrated on the convergence rates of global and personalized models under various heterogeneity constraints.  While some work has sought to improve the performance of one or both scenarios, .  These efforts each have their own limitations: a global model shared by all peers does not provide privacy, while personalized models are catered to an individuals data distribution.  In this work.  
\label{discussion}

\section*{Acknowledgement}
This work was  supported  in  part  by NSF under award numbers ATD 2123761, CNS 2335687, CNS 1822118 and from NIST, Statnett, Newpush, Cyber Risk Research, AMI, and ARL.
%\section*{References}

\vspace{-0.7em}
%\begin{thebibliography}{00}
%\bibliography{ipa}
%\end{thebibliography}
\bibliographystyle{IEEEtran}
\bibliography{ipa}

%\appendix
%\section*{Appendix}

\end{document}